\documentclass[10pt,twocolumn,letterpaper]{article}

\usepackage{cvpr}              %

\definecolor{cvprblue}{rgb}{0.21,0.49,0.74}
\usepackage[pagebackref,breaklinks,colorlinks,allcolors=cvprblue]{hyperref}
\usepackage{graphicx}
\usepackage{float}
\usepackage{amsmath}
\usepackage{multicol}
\usepackage{multirow}
\usepackage{pifont}
\usepackage{booktabs}
\usepackage{makecell}
\usepackage{enumitem}
\usepackage{stfloats}
\usepackage{placeins}
\usepackage{caption}

\newcommand{\method}{CaptionFormer}
\newcommand{\idata}{LVISCap}
\newcommand{\vdata}{LV-VISCap}

\newcommand{\gf}[1]{\textcolor{black}{#1}}
\newcommand{\gff}[1]{\textcolor{black}{#1}}

\newcommand{\mdf}[1]{\textcolor{black}{#1}}

\newcommand{\camver}[1]{\textcolor{black}{#1}}

\definecolor{lightgray}{gray}{0.40}  %

\def\paperID{} %
\def\confName{CVPR}
\def\confYear{2026}

\title{\method: Unified Segmentation, Tracking, and Captioning for Spatio-Temporal Objects}

\author{
Gabriel Fiastre$^{1*}$ \quad\quad\quad
Antoine Yang$^{2}$ \quad\quad\quad
Cordelia Schmid$^{1}$\\
{
$^{1}$Inria, École Normale Supérieure, CNRS, PSL Research University \quad
$^{2}$Google DeepMind
} \\
\href{https://www.gabriel.fiastre.fr/captionformer/}{\texttt{https://www.gabriel.fiastre.fr/captionformer/}}
}

\begin{document}

\maketitle

\begingroup
\renewcommand\thefootnote{*}
\footnotetext{Corresponding author: \texttt{gabriel.fiastre@inria.fr}}
\endgroup

\begin{abstract}

Dense Video Object Captioning (DVOC) is the task of jointly detecting, tracking, and captioning object trajectories in a video, requiring the ability to understand spatio-temporal details and describe them in natural language.
Due to the complexity of the task and the high cost associated with manual annotation, 
previous approaches resort to training strategies with limited data, potentially leading to suboptimal performance.
To circumvent this issue,  we propose to generate captions about spatio-temporally localized entities leveraging a state-of-the-art VLM, and extend the LVIS and LV-VIS datasets with our synthetic captions (LVISCap and LV-VISCap). Moreover, we introduce an end-to-end model, \method{}, capable of jointly detecting, segmenting, tracking and captioning object trajectories.
\mdf{With pretraining on LVISCap and LV-VISCap,} \method{} achieves state-of-the-art DVOC results on three existing benchmarks, VidSTG, VLN and BenSMOT. 

\end{abstract}

\vspace{-0.3em}
\section{Introduction}
A fundamental aim of computer vision is to enable machines to understand videos with human-like acuity in perceiving and reasoning about the world. 
Recent advances have led to remarkable progress in both spatio-temporal localization \citep{NIPS2015_14bfa6bb,Redmon_2016_CVPR,carion2020end,he2017mask,wojke2017simple,zhang2022bytetrack} and vision-language understanding \citep{vinyals2015show, sun2019videobert, yu2019deep, yang2021just, chen2020uniter, jia2021scaling, zellers2021merlot, bain2021frozen, lu2019vilbert,mdetr}.
However, building vision-language models that can simultaneously reason about spatially localized objects and temporal dynamics of a complex scene remains a significant challenge, motivated by many real-world applications including autonomous driving \citep{kim2019grounding, atakishiyev2024explainable}, human-computer interaction \citep{shridhar2020alfred, ahn2022can}, or video editing  \citep{molad2023dreamix,jeong2023ground}.
Dense Video Object Captioning (DVOC) \citep{zhou2023dense} serves as a key benchmark for this purpose, as it requires to jointly \camver{detect}, track, and describe in natural language \emph{all} visual entities in a video.

\begin{figure}[htb!]
    \centering
    \includegraphics[width=\linewidth]{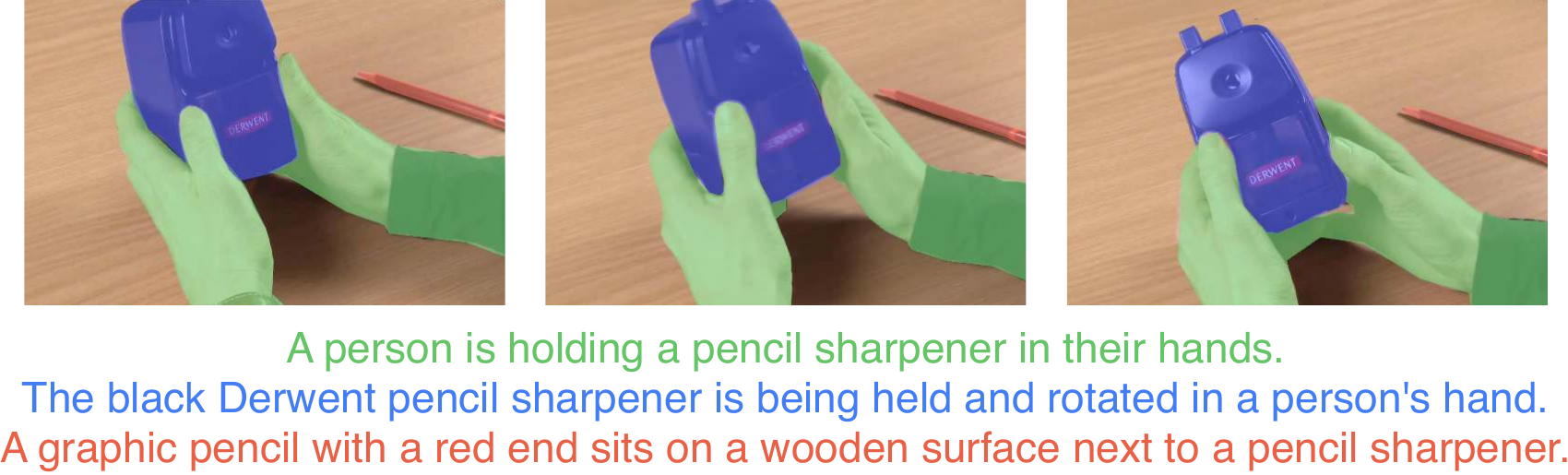}
    \vspace{-1.7em}
    \caption{{Example of synthetic captions in our \vdata{} dataset.}
    }
    \label{fig:teaser}
    \vspace{-1.5em}
\end{figure}

Manual annotation for such a fine-grained task is particularly expensive, leading to a scarcity of datasets with densely annotated object-level video descriptions.
To tackle DVOC, prior work resorted to alternative training approaches:
\citet{zhou2023dense} propose a disjoint training strategy, decomposing the problem into subtasks and training a model sequentially on %
datasets for each subtask.
\citet{choudhuriow} leverage the pretraining of multiple specialized models to alleviate the need for object-level annotations. 
Both methods allowed to perform DVOC while circumventing the need for costly annotations, but the lack of end-to-end training with densely annotated object-level supervision may lead to %
suboptimal performance.

We propose to address this limitation by generating synthetic object-level annotations, motivated by the recent success of LLM-generated supervision \citep{liu2023visual, abdin2024phi} and the growing visual capacities of Vision Language Models (VLMs) \citep{alayrac2022flamingo, li2022blip, li2023blip, team2023gemini, team2024gemini, achiam2023gpt, grattafiori2024llama,bai2023qwen}. 
To the best of our knowledge, our work is the first to generate localized, object-level captions \mdf{for DVOC}. 
To this end, we introduce a multi-modal prompting strategy 
leveraging a state-of-the-art VLM, and extend two segmentation datasets, LVIS \citep{gupta2019lvis} for images and LV-VIS \citep{Wang_2023_ICCV} for videos, to be the first DVOC training sets with (mask, box, category, caption) annotations for all objects, dubbed \idata{} and \vdata{}, see figure~\ref{fig:teaser}. 

Using our generated datasets, we extend the DVOC task \camver{from detection to segmentation} and train \method{}, the first end-to-end model \camver{to} jointly produce (mask, caption) pairs for all object trajectories in a video. 
We show that 
(i)~our generated datasets, \idata{} and \vdata{}, largely benefit \method{}'s DVOC performance,
(ii)~our \method{} outperforms previous state-of-the-art
on the VidSTG, VLN and BenSMOT 
benchmarks and \mdf{(iii)~we can extend \camver{DVOC} to segmentation}. \\
Overall, our contributions can be summarized as follows: 
\begin{itemize}[leftmargin=0.5cm,topsep=0pt,noitemsep]
    \item[1.] We introduce a VLM-based method to generate synthetic object captions for videos, and extend the LVIS and LV-VIS datasets to be the first unified DVOC training set with object captions, boxes, and segmentation masks: \idata{} and \vdata{}. 
    \item[2.] Using our unified generated data, we train \method{}, the first end-to-end model to jointly detect, segment, track and caption objects in a video. 
    \item[3.]  %
    \method{} achieves state-of-the-art DVOC results on the three existing benchmarks : VidSTG, VLN and BenSMOT.
\end{itemize}
\camver{The code and generated annotations are available at \citep{webpage}}.

\section{Related work}

\noindent\textbf{Open-Vocabulary Video Instance Segmentation (OV-VIS).} The OV-VIS task aims to segment, track, and classify objects from an open set of categories in videos \citep{guo2025openvis, Wang_2023_ICCV}, using datasets such as LV-VIS \citep{Wang_2023_ICCV}.
State-of-the-art methods \citep{guo2025openvis, Wang_2023_ICCV,fang2024unified} commonly use query-based approaches that classify objects by matching visual features with CLIP embeddings \citep{radford2021learning}. 
Methods like OVFormer \citep{fang2024unified} or BriVIS \citep{cheng2024instance} improve this approach by better aligning visual queries with the CLIP space. 
Unlike these methods focused on CLIP feature matching for classification, our work explores describing objects in natural language \citep{li2023blip}.

\noindent\textbf{Localized vision-language understanding.} 
Going beyond pioneering vision-language tasks such as visual question answering \citep{antol2015vqa} or image captioning \citep{chen2015microsoft}, recent work has explored spatial understanding tasks that require localizing natural language queries in images.
This includes referred expression segmentation \citep{kazemzadeh2014referitgame, yu2018mattnet, yang2022lavt}, image grounding \citep{rohrbach2016grounding, plummer2015flickr30k}, reasoning segmentation \citep{lai2024lisa,wang2024llm}, spatio-temporal video grounding \citep{zhang2020does, Yang_2022_CVPR} and grounded visual question answering \citep{zhu2016visual7w,xiao2024can, lei2018tvqa, lei2019tvqa+}.
While these tasks typically require localizing one or a few entities, dense captioning \citep{johnson2016densecap, wu2024grit} aims to spatially localize and describe in natural language \emph{all} salient regions in images.
Our work addresses the more challenging task of predicting \emph{both} object trajectories and descriptions for \emph{all} objects in a video.

\noindent\textbf{Dense Video Object Captioning (DVOC).} The DVOC task aims at jointly detecting, tracking, and describing the trajectory of all visual entities in a video. 
DVOC-DS \citep{zhou2023dense} tackles this task by generating frame-wise object box proposals \citep{zhou2019objects, cai2018cascade} that are tracked \citep{zhou2022global} before feeding aggregated and cropped features to a generative image-to-text decoder \citep{wang2022git}.
To cope with the lack of DVOC annotations, the model is trained disjointly on various subtasks: object detection using COCO \citep{lin2014microsoft}, image object-level captioning using Visual Genome \citep{krishna2017visual}, video scene-level captioning using SMiT \citep{monfort2021spoken} and video object tracking using AugCOCO \citep{lin2014microsoft}.
OW-VisCapTor~\citep{choudhuriow} extends the DVOC task to segmentation masks. 
Its architecture relies on an object abstractor using a prompt encoder and transformer blocks, an inter-query contrastive loss to ensure object queries are diverse, and an object-to-text abstractor that connects these queries to a frozen LLM, generating rich, object-centric captions for each detected instance.
While proposing to extend DVOC to segmentation, OW-VisCapTor is hindered by the absence of paired (mask, caption) annotations, hence segmentation and DVOC are tackled in isolation using separate models. 
Closely related, SMOTer \citep{li2024beyond} extends multi-object tracking to object-level captioning, video-level captioning and object-relation predictions, and introduce a hand-annotated dataset, BenSMOT. 
However, BenSMOT focuses on humans only, whereas the standard DVOC task considers all visual entities in the video.
In this work, we automatically generate DVOC datasets, \idata{} and \vdata{}, and train \method{}, a model that can end-to-end detect, segment, track and caption object trajectories in videos.

\begin{figure*}[tbp]
    \centering
    \includegraphics[width=\linewidth]{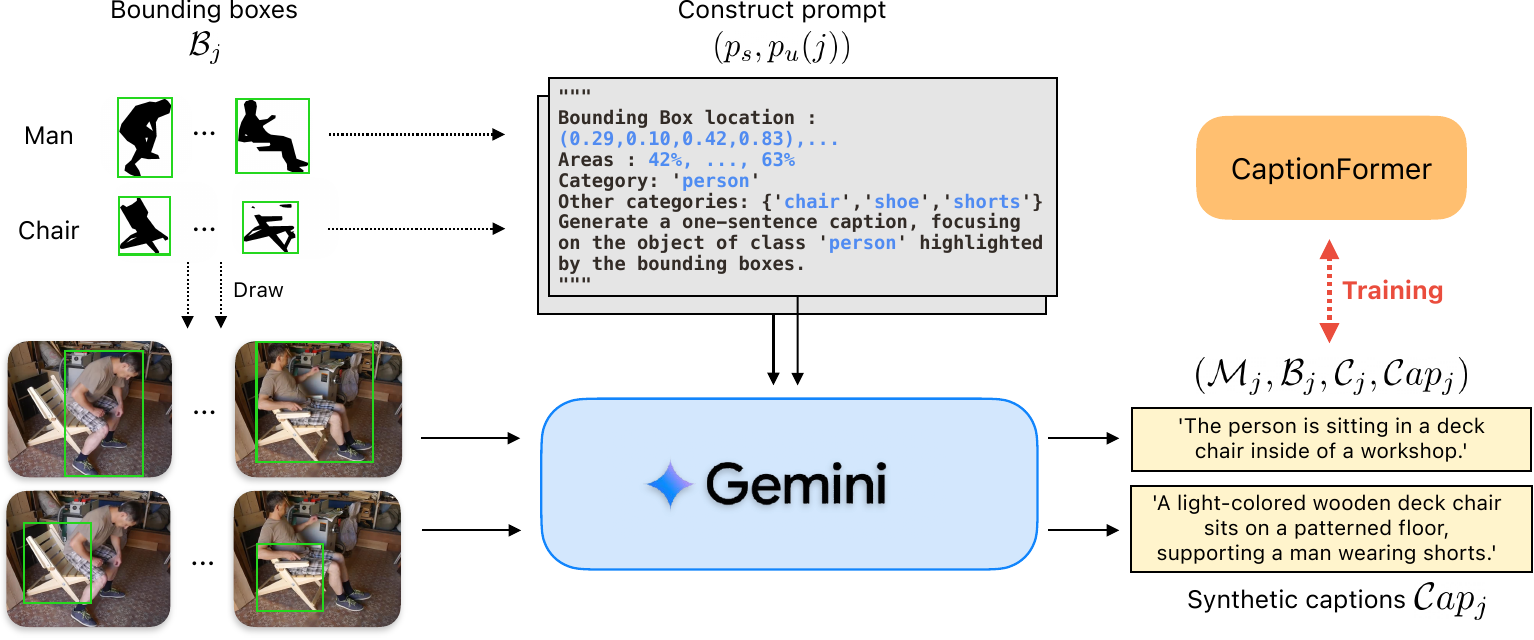}
    \caption{\textbf{Our \method{} data annotation pipeline}:
    For each object, we extract its bounding boxes $\mathcal{B}_j$ from its annotated masks $\mathcal{M}_j$, and draw them in the video.
    The video with drawn bounding boxes $\hat{x}_j$, along with a prompt ($p_s, p_u(j)$) including additional information such as the label of the object to caption $\mathcal{C}_j$, the labels and bounding boxes of other objects in the image, is fed to a state-of-the-art VLM (Gemini 2.0 Flash \citep{team2023gemini}) to generate the object-level caption. 
    $p_s$ denotes the static prompt with general instructions and $p_u(j)$ the dynamic prompt with annotation information.
    We use this pipeline to generate the \idata{} and \vdata{} datasets, used to train \method{}.
    }
    \label{fig:annotation_loop}
    \vspace{-0.5cm}
\end{figure*}

\noindent\textbf{Vision-language data generation.}
A promising approach for intricate vision-language tasks is to generate visual annotations using Large Language Models (LLMs) or Vision Language Models (VLMs).
LLaVA \citep{liu2023visual} leverages a LLM to generate conversations and detailed descriptions from paired image-text annotations~\citep{chen2015microsoft}.
This approach has been followed to generate large-scale instruction tuning data for video understanding \citep{maaz2023video, li2023videochat}.
Recent research has focused on the generation of spatially grounded text-image data using LLMs/VLMs: 
Shikra \citep{chen2023shikra} generates grounded QA pairs from Flickr30K Entities \citep{plummer2015flickr30k}, LLaVA-Grounding \citep{zhang2024llava} and GLaMM \citep{rasheed2024glamm} generate grounded conversational data from datasets such as COCO \citep{lin2014microsoft}.
Recently, GROVE \citep{kazakos2025large} extended GLaMM to generate grounded, temporally consistent video captions, extending dense captioning to the video domain.
These methods predominantly operate at the scene level and do not produce localized, object-level video descriptions. \mdf{In contrast, we introduce a multi-modal prompting strategy to leverage VLMs into generating fine-grained captions for individual object trajectories across time. Recently, \cite{sa2va} proposed a related VLM-based approach to generate object-level video captions designed for long sentences referring segmentation. Differently, we focus on the task of Dense Video Object Captioning where no text prompt is given to the model.}

\section{Method}\label{sec:method}
Dense Video Object Captioning (DVOC) \citep{zhou2023dense} is the task of jointly detecting, tracking and captioning objects trajectories in a video, i.e. producing a (bounding box or mask, caption) pair for each object in a video at each timestamp they appear. 
Such densely annotated data is lacking for video, as there is currently no available training dataset including captions for all object trajectories in the videos.
The spatio-temporal grounding dataset VidSTG \citep{zhang2020does} has been repurposed to DVOC but only includes annotations for few objects per video and a limited number of frames per video.

We address the lack of data by introducing a strategy for synthetic DVOC data generation, allowing unified training of a DVOC model on (trajectory, caption) pairs for each object, as presented in Section \ref{sec:datagen}.
With this strategy, we extend the LV-VIS dataset \citep{Wang_2023_ICCV}, which includes both boxes and masks in their (trajectory, label) annotations for all objects, to a variant, dubbed \vdata{}, which additionally includes synthetically generated captions for each trajectory.
To enable end-to-end training on this rich data using segmentation masks, we build an architecture based on a state-of-the-art Open-Vocabulary Video Instance Segmentation (OV-VIS) model OVFormer \citep{fang2024unified}, as described in Section \ref{sec:clipmodel}.
As OVFormer is designed for classification, we extend it with a captioning head \citep{choudhuriow}, as explained in Section \ref{sec:videomodel}.
Finally, we present in Section \ref{sec:training} the losses used to train our model.

\vspace{-0.1cm}
\subsection{DVOC data generation}\label{sec:datagen}
We start from the LV-VIS dataset \citep{Wang_2023_ICCV} which contains (segmentation masks, category) manual annotations for all objects and timestamps of the videos.
To automatically collect DVOC data, one challenge is to generate accurate object-level captions for each trajectory.
For this, we leverage a state-of-the-art VLM (Gemini 2.0 Flash \citep{team2023gemini}) and feed it with videos where the object to caption is marked with drawn bounding boxes, as illustrated in Figure \ref{fig:annotation_loop}.

\begin{table}
\caption{\textbf{Impact of the prompting strategy on caption quality}. Scores are given by an expert human evaluator from 0 to 2 (incorrect, partially correct, or correct) on a subset from the LV-VIS validation set, and \camver{scaled} to 0-100 range. For the mask visual prompt experiments, we use our best prompt with either the object's bounding boxes or center point coordinates as a localization cue in the text prompt.}\label{tab:prompt-ablation}
\centering
\resizebox{0.47\textwidth}{!}{
\begin{tabular}{c|l|c}
\toprule
\textbf{\makecell[c]{Visual prompt}} & \textbf{\makecell[c]{Prompting method}} & \textbf{\makecell[c]{Avg rating}} \\
\midrule
\multirow{7}{*}{\makecell{bounding boxes}} & single frame & 26.8 \\
& + multiple frames & 27.1 \\
& + detailed instructions & 29.5 \\
& + category labels & 80.7 \\
& + bbox coordinates & 83.1 \\
& + bbox area & \underline{84.3} \\
& + few shot examples & \textbf{85.1} \\
\midrule
\multirow{2}{*}{\makecell{mask boundaries}} & center point coordinates & 75.9 \\
& bbox coordinates & 77.1 \\
\bottomrule
\end{tabular}}
\vspace{-1em}
\end{table}

\noindent \textbf{Visual Prompt.}
Formally, let $x\in \mathbb{R}^{N\times H \times W\times 3}$ be a video clip of length $N$, with associated mask and category annotations for $M$ objects in the video: $(\mathcal{M}_j, \mathcal{C}at_j),\ j\in {1,..M}$. 
We first extract bounding boxes annotations from the ground-truth masks, $\mathcal{B}_j\in\mathbb{R}^{ N\times4},\ j\in1,...,M$.
We draw each object boxes on a separate copy of the video, and denote as $a\wedge b$ the operation of drawing box $b$ on frame $a$. 
We obtain a visual prompt $\hat{x}_j^i$ for each object trajectory: 
\begin{equation}
    \hat{x}_j^i = x^i \wedge\mathcal{B}_j\ \textrm{ for}\ i\in 1,...,N,\ j \in 1,...,M
\end{equation}

Note that in practice, we subsample $N$ to 4 uniformly sampled video frames as we found it produces representative enough visual content for the captioning task.

\begin{figure*}[t!]
    \centering
    \includegraphics[width=0.95\linewidth]{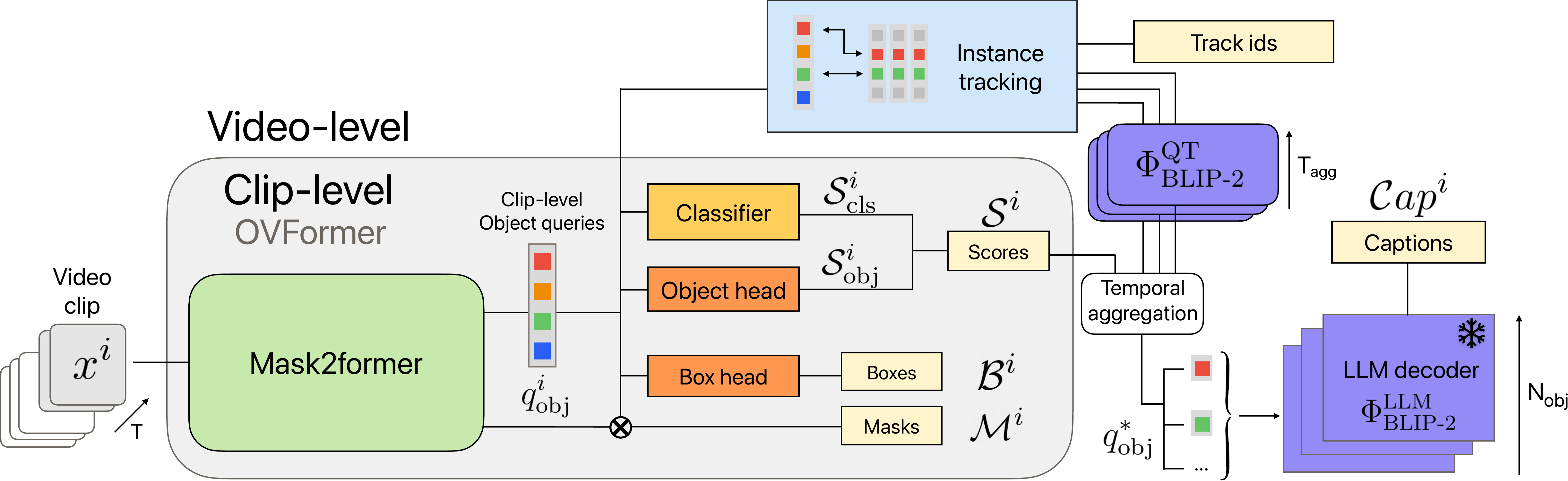}
    \caption{\textbf{Our \method{} architecture} jointly segments and captions objects in videos. 
    For each clip of $T$ frames, we obtain $N_{\textrm{obj}}$ clip-level object queries through Mask2Former \citep{cheng2021mask2formervideoinstancesegmentation}, and yield associated score, mask and box predictions. 
    At a video-level, we match the predicted object queries with the previously processed clips with the Hungarian bipartite matching algorithm, and perform instance tracking~\citep{zhu2024clip}. 
    \mdf{For each track, we sample $T_{\textrm{agg}}$ clips uniformly across the video, and aggregate the tracked queries to obtain a single video query per object, which we can feed to the LLM captioning head \citep{li2023blip}, producing a single caption per trajectory.}}
    \label{fig:architecture}
    \vspace{-0.7em}
\end{figure*}

\noindent \textbf{Text prompts.}
In detail, the prompt we feed the VLM is composed of the previously described visual prompt $\hat{x}_j$, a system prompt $p_s$ and a user prompt $p_u(j)$. 
The system prompt is static for all objects/videos and gives general instructions (e.g.~"Generate a caption about a queried object highlighted with bounding boxes."), rules (e.g.~"Do not mention the bounding boxes in the caption."), format, and an example. 
The user prompt $p_u(j)$ however, is constructed for each annotation and enriched with informations about the specific query to help the model. 
The user prompt encodes textual bounding box coordinates, area, category name of the queried object, and category names from other objects in the image. 
Passing information through different channels (visual prompt, text prompt) helps the model focusing on the queried object and being accurate when describing the scene. 
Also, this complementary information can lead the model to reason about the objects (e.g.~area being small for an object of category 'elephant' implies the object is most likely part of the background). 
\mdf{The different prompting cues have been ablated in Table \ref{tab:prompt-ablation}. In particular, showing that including textual semantic and localization cues in the prompt helps the model to focus on the queried object and generate more accurate object-focused captions. We notice that using the segmentation masks as the visual prompt for the model results in less accurate object captions, which might result from a poorer alignment with the localization cues included in the text prompt.} The full template details for the model prompt construction and a more detailed ablation are given in the supplementary material~\ref{app-subsec:prompt-strat} and~\ref{app-sec:ablations}.

\noindent \textbf{The \idata{} and \vdata{} datasets.}
For each object $j$ in the video, we prompt Gemini 2.0 Flash \citep{team2023gemini} with the visual and textual prompts ($\hat{x}_j$ , $p_s$ and $p_u(j)$) to output a synthetic caption $\mathcal{C}ap_j$.
The full DVOC annotation is $(\mathcal{M}_j, \mathcal{B}_j, \mathcal{C}at_j, \mathcal{C}ap_j)$. 
We repeat this process for each object in each videos of LV-VIS, generating over $19.5$k synthetic captions over $3.9$k videos covering $1,020$ object classes, and with an average length of $13.9$ words.
We repeat this process with the LVIS dataset to support image pre-training, applying the same method as above described with each image considered a video of len $N=1$, and obtain our DVOC training sets: \idata{} and \vdata{}.

\subsection{\method{} architecture}\label{sec:model}

\mdf{To enable end-to-end training on the previously described DVOC data including segmentation masks, our architecture, illustrated in Figure \ref{fig:architecture}, processes videos as $N_{\textrm{clip}}$ clips of $T$ consecutive frames each, and is composed with:}
(i) at the clip level, an instance segmentation and detection component, see Section \ref{sec:clipmodel}
(ii) at the video-level, a tracking module and a captioning head, see Section \ref{sec:videomodel}.
\subsubsection{Instance segmentation and detection}\label{sec:clipmodel}

\noindent \textbf{Background: OVFormer~\citep{fang2024unified}.}
Our clip-level instance segmentation component is based on OVFormer~\citep{fang2024unified}, which we shortly describe next.
On a high level, OVFormer augments Mask2Former \citep{cheng2021mask2formervideoinstancesegmentation} with a classification head to handle Open-Vocabulary Video Instance Segmentation.
Mask2Former learns clip-level object queries with a transformer decoder which cross-attends to the features extracted with a visual backbone and are refined with a pixel decoder.
Formally, each clip $x^i_{\textrm{clip}}\in \mathbb{R}^{T\times 3\times H\times W}$ is composed of $T$ consecutive frames for images of resolution ($H$,$W$), and processed independently by the OVFormer model which outputs clip-level object queries $q_{\textrm{obj}}^i$, associated mask predictions $\mathcal{M}^i$, classification and objectness scores $\mathcal{S}_{\textrm{cls}}^i$ and $\mathcal{S}_{\textrm{obj}}^i$: 
$(q_{\textrm{obj}}, \mathcal{M}^i, \mathcal{S}_{\textrm{cls}}^i, \mathcal{S}_{\textrm{obj}}^i) = \Phi_{\textrm{OVFormer}}(x^i)$
where $q_{\textrm{obj}}^i\in \mathbb{R}^{N_{\textrm{obj}}\times D}$, $\mathcal{M}^i\in\mathbb{R}^{N_{\textrm{obj}}\times T\times H \times W}$, $S^i_{\textrm{cls}}\in\mathbb{R}^{N_{\textrm{clip}}\times N_{\textrm{obj}}\times N_{\textrm{cls}}}$ and $S^i_{\textrm{obj}}\in\mathbb{R}^{N_{\textrm{clip}}\times N_{\textrm{obj}}}$.

\clearpage
\noindent \textbf{Detection head.}
We extend this segmentation module with detection by using a 4-layer MLP to generate boxes on top of the object queries $q^i_{\textrm{obj}}$:  $ \mathcal{B}^i=\textrm{BoxHead}(q^i_{\textrm{obj}})\in\mathbb{R}^{N_{\textrm{obj}}\times T \times 4}$.

\noindent \textbf{Confidence scores.}
Note that OVFormer \citep{fang2024unified} only computes class-aware query-wise confidence scores over the full video. 
However, for objects appearing only in a small subset of frames in the video this strategy could result in inaccurate scores.
Moreover, for DVOC, we wish to avoid redundant predictions i.e. having two queries predicting a similar trajectory.
Hence we additionally compute \emph{class-agnostic} query-wise confidence scores \emph{for each clip} $\mathcal{S}_{\textrm{cls}*}^i$, by taking the maximum classification score over all labels $c\in 1,...,N_{\textrm{cls}}$ for each query and clip: \\
\begin{equation}
    \mathcal{S}_{\textrm{cls}*}^i = \underset{c}{max}(S_{\textrm{cls}}^i(c)) \in \mathbb{R}^{N_{\textrm{clip}}\times N_{\textrm{obj}}}
\end{equation}
\vspace{-0.2em}
Finally we derive the per-clip score $\mathcal{S}^i=\sqrt{\mathbf{S}^i_{\textrm{cls}*}\times \mathbf{S}^i_{\textrm{obj}}}$ which we use for filtering predictions below a threshold $t_{\textrm{thresh}}$ at inference-time for every time step.

\subsubsection{Instance tracking and captioning}\label{sec:videomodel}
\noindent \textbf{Tracking module.}
To derive the output video-level trajectories from the clip-level predictions, we need to obtain a matching between the queries at time $i$ and the queries at time $i+1$. 
For this, we perform tracking between the clips using the top-K enhanced query-matching module from \citet{zhu2024clip}. 
For each clip, this module keeps a memory bank containing the queries from the $T_{\textrm{match}}$ previous clips. 
Among these, it identifies the $K_{\textrm{match}}$ most matched clips, and computes the optimal assignment using the Hungarian bipartite matching algorithm.
Using the $K_{\textrm{match}}$ most matched clips helps reducing error propagation compared to the OVFormer \citep{fang2024unified} tracking module, which maps queries from time $i+1$ to time $i$ directly. 
Notably, we can keep track of objects that disappear and re-appear in a video, whereas they are automatically lost using the OVFormer tracking module. 

This method is referred to as semi-online tracking as we represent objects at a clip-level and associate between the clips in an online fashion. This offers the advantages of being flexible 
(fully-online for clips of length $T=1$) and to arbitrate between using multi-frame information and memory constraints for long videos.

\noindent \textbf{Captioning head.}
To caption tracked object trajectories, we adapt the captioning head from \citet{choudhuriow} based on BLIP-2 \citep{li2023blip}.
The BLIP-2 decoder processes object queries one by one using masked-attention conditioned with the predicted masks, before projecting the resulting object query into the LLM space for caption prediction.
However, for consistency and efficiency, we predict a single video-level caption per tracked object query, replacing clip-level prediction.

\noindent \mdf{\textbf{Query aggregation.} \gf{Unlike previous works \cite{choudhuriow, li2023blip}, we extend the BLIP-2 decoder with temporal aggregation.} Let's consider $\Phi^{\textrm{QT}}_{\textrm{BLIP2}}$ the BLIP-2 query transformer, $\Phi^{\textrm{LLM}}_{\textrm{BLIP2}}$ the LLM decoder, ${q}_{\textrm{obj}}^i(j)\in \mathbb{R}^{1\times D}$ and $\mathcal{S}_{\textrm{obj}}^i(j)\in \mathbb{R}$ respectively the query and the detection score for object $j$ from clip $i$. 
For each object, we aggregate the tracked queries over time after they are processed by the BLIP-2 query transformer, by sampling a set $\mathcal{I_{\textrm{agg}}}$ of  $T_{\textrm{agg}}$ clips uniformly across the video. We obtain a video query for each object $j$:} 
\begin{equation}
    q^*_{\textrm{obj}}(j)=\sum_{i\in{\mathcal{I_{\textrm{agg}}}}}{\mathcal{S}^i(j)\times \Phi^{\textrm{QT}}_{\textrm{BLIP-2}}(q_{\textrm{obj}}^i(j), \mathcal{M}^{i}_j)}
\end{equation}
\mdf{\gf{We then compute the video captioning prediction of tracked object $j$ from its aggregated query:}} 
\begin{equation}
    \mathcal{C}ap(j) = \Phi_{\textrm{BLIP2}}^{\textrm{LLM}}(q_{\textrm{obj}}^{*}(j)) \textrm{ for } j\in1,...N
\end{equation}
\gff{The impact of using temporal aggregation is detailed in Section \ref{sec:results}, and the choice of aggregation strategy is shown to be important in Section \ref{sec:additional}.}

\subsection{Model training}\label{sec:training}

We train \method{} \mdf{with a combination of clip-level and video-level losses}. For each clip, we predict masks/boxes, classification scores and captions, and derive the clip-level training objective as the following combination of supervised losses:
\begin{equation}
\mathcal{L}_{\textrm{clip-level}}=\mathcal{L}_{\textrm{seg}}+\mathcal{L}_{\textrm{det}} + \mathcal{L}_{\textrm{s}} + \mathcal{L}_{\textrm{cap}}
\end{equation}
where $\mathcal{L}_{\textrm{seg}}$ and $\mathcal{L}_{\textrm{s}}$ are the VIS losses from \citet{fang2024unified}, i.e. $
     \mathcal{L}_{\textrm{seg}}= \lambda_{\textrm{dice}}\mathcal{L}_{\textrm{dice}} + \lambda_{\textrm{ce}}\mathcal{L}_{\textrm{ce}} $,
with $\mathcal{L}_{\textrm{dice}}$ and $\mathcal{L}_{\textrm{ce}}$ the dice and cross-entropy segmentation losses respectively, and $\mathcal{L}_s= \lambda_{\textrm{cls}}\mathcal{L}_{\textrm{cls}} + \lambda_{\textrm{obj}}\mathcal{L}_{\textrm{obj}}$ where $\mathcal{L}_{\textrm{cls}}$ and $\mathcal{L}_{\textrm{obj}}$ are the cross-entropy losses for classification and objectness.
We add detection and captioning losses $\mathcal{L}_{\textrm{det}}=\lambda_{l_1}\mathcal{L}_{l_1} + \lambda_{\textrm{giou}}\mathcal{L}_{\textrm{giou}}$ where $\mathcal{L}_{l_1}$ and $\mathcal{L}_{\textrm{giou}}$ are detection losses from \citet{Yang_2022_CVPR}, and $\mathcal{L}_{\textrm{cap}}=\lambda_{\textrm{clip-lm}}\mathcal{L}_{\textrm{lm}}$, with $\mathcal{L}_{\textrm{lm}}$ the cross-entropy language modeling loss \citep{zhou2023dense}.

\mdf{When including the temporal aggregation module for captioning, we train the captioning head at the video level, i.e. we predict a caption per object for the full video after the tracking is performed and each object-query has been augmented across time.}
\vspace{-0.2em}
\begin{equation}
    \mathcal{L}_{\textrm{video-level}}=\lambda_{\textrm{vid-lm}}\mathcal{L}_{\textrm{lm}}
\end{equation}
\vspace{-1em}

\mdf{\method{} can be trained in a completely end-to-end manner. However, in practice, we train the model in two-stages for most of the experiments to alleviate memory constraints: we first train the segmentation/detection  and classification model, then freeze it and tune the captioning head. The captioning head is trained either at the clip-level, or at the video-level when enabling the temporal aggregation module (i.e. $\lambda_{\textrm{clip-lm}}=0$ or $\lambda_{\textrm{vid-lm}}=0$). }
We set each loss respective weight when computed to
$\lambda_{\textrm{dice}}, \lambda_{\textrm{ce}}, \lambda_{l_1}=5$, $\lambda_{\textrm{giou}}, \lambda_{\textrm{cls}}, \lambda_{\textrm{obj}}=2$, and ($\lambda_{\textrm{clip-lm}}=1$ or $\lambda_{\textrm{vid-lm}}=1$).

\section{Experiments}

\begin{table}[t!]
\centering
\caption{\textbf{Impact of training with \idata{} and \vdata{} and of the visual backbone on \vdata{} DVOC. }}
\resizebox{0.48\textwidth}{!}{
\begin{tabular}{c|cc|ccc|c}
\toprule
\textbf{Backbone} & \textbf{LVIScap} & \textbf{LV-VIScap} & \textbf{CapA} & \textbf{DetA} & \textbf{AssA} & \textbf{CHOTA} \\

\toprule
 \multirow{3}*{\textbf{SwinB}} & - & \ding{51} & 37.9 & 48.1 & 89.5 & 54.7 \\
 & \ding{51} & - & 30.0 & 34.3 & \textbf{93.2} & 45.8 \\
 
 &\ding{51} & \ding{51} & \textbf{43.6} & \textbf{54.3} & 89.1 & \textbf{59.5} \\
 \midrule
 \textbf{ResNet50}& \ding{51} & \ding{51} & {39.0} & {51.1} & {88.5} & {56.1}  \\
\bottomrule
\end{tabular}}
\label{tab:lvviscap}
\vspace{-1em}
\end{table}

\subsection{Experimental setting}\label{sec:datasets}
\textbf{Datasets.}
\textbf{LVIS} \citep{gupta2019lvis} and 
\textbf{LV-VIS}\citep{Wang_2023_ICCV} are large-vocabulary instance segmentation datasets, respectively for image and video. \textbf{\idata{}} and \textbf{\vdata{}} denote our extensions of LVIS and LV-VIS (see Section \ref{sec:datagen}), with respectively $1,2M/244k$ synthetic image object captions for the training/validation of \idata{} and $16k/3.7k$ synthetic video object captions for \vdata{} (average of $5.4$ objects per video). Note that in the absence of annotations on the test sets of LVIS and LV-VIS, we only extend the training and validation sets with captions, and use the validation set for evaluation.

\noindent \textbf{Benchmarks.} \textbf{VidSTG}\citep{zhang2020does} is a spatio-temporal video grounding dataset containing text descriptions serving as queries, which \citet{zhou2023dense} propose to use for DVOC evaluation. The repurposed training and validation sets count $5.4k/602$ videos for $15.1k/1.6k$ object trajectories with captions respectively. \textbf{Video Localized Narratives (VLN)} is similarly repurposed \citep{zhou2023dense}. For each of the $5.1k$ training and $2.4k$ validation videos, the dataset contains 3 sparsely annotated frames with non exhaustive captions. \textbf{BenSMOT} contains bounding box trajectories and associated captions focusing exclusively on humans in videos, with an average of $2.2$ instances per video. It counts $2.2k$ videos for training and $1k$ for evaluation. 
More details about the datasets are given in the supplementary material ~\ref{app-sec:datasets}.

\noindent \textbf{Evaluation Metrics}. 
Following prior work, we evaluate DVOC using the CHOTA metric introduced by \citet{zhou2023dense}, which extends the widely used multi-object tracking HOTA metric \citep{luiten2021hota}. 
CHOTA decomposes the DVOC task into three components: detection accuracy (DetA)\citep{luiten2021hota}, association accuracy (AssA)\citep{luiten2021hota}, and captioning accuracy (CapA)\citep{zhou2023dense}. 
These components reflect the model's ability to 
(i) correctly localize objects, 
(ii) maintain their identity across frames, and 
(iii) generate accurate natural language descriptions. 
CHOTA is defined as the geometric mean of the three components: $\textrm{CHOTA}=\sqrt[3]{\textrm{DetA}\cdot \textrm{AssA}\cdot \textrm{CapA}}$.
Matching between predicted and ground-truth trajectories is performed using Intersection-over-Union (IoU) thresholds, similarly to standard tracking evaluations \citep{milan2016mot16}. 
To extend the metric to segmentation masks, we simply replace box-based IoU with mask-based IoU when computing CHOTA with segmentation masks.

\noindent \textbf{Implementation details.}
Following OVFormer \citep{fang2024unified} we use ResNet50 \citep{he2016deep} and SwinBase \citep{liu2021swin} visual backbones; 
our Mask2former \citep{Cheng_2022_CVPR} transformer decoder has $11$ layers, and our captioning head is based on the BLIP-2 \citep{li2023blip} decoder with OPT-2.7B LLM \citep{zhang2022opt}. 
\mdf{For LV-VIS experiments we tune the model end-to-end with clip-level supervision only. For other experiments, we first train the segmentation/detection model, then freeze it and tune the captioning head.  For VidSTG, VLN and BenSMOT experiments we use video-level tuning for captioning with temporal aggregation.}
For the largest dataset (COCO + LVIS) the optimization takes 2 days on 4 H100 GPUs.
More implementation details and hyper-parameters for different experiments are given in the supplementary material~\ref{app-sec:implementation}.

\begin{figure}
    \centering
    \includegraphics[width=0.9\linewidth]{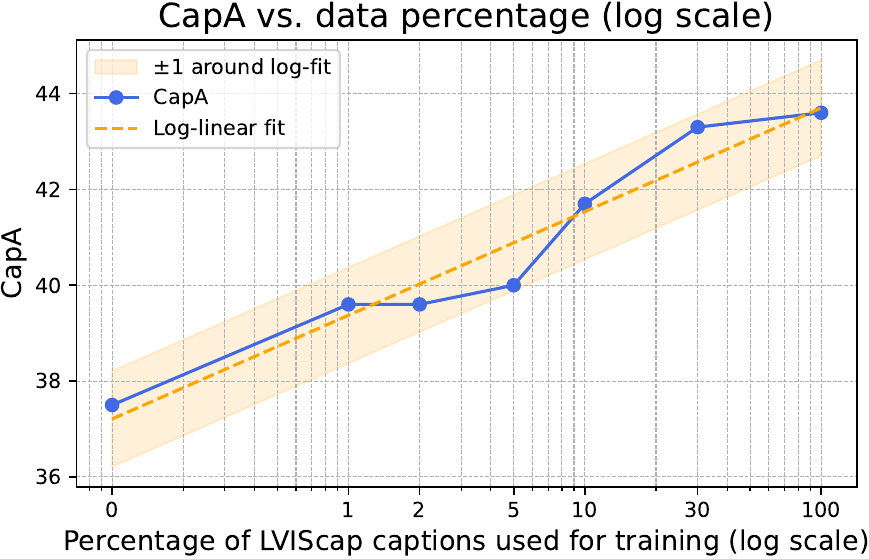}
    \caption{\mdf{\textbf{Impact of the generated data scale on the CapA metric.} 
    We train \method{} on a varying percentage of \idata{} captions and finetune on \vdata{}.}}
    \label{fig:scaling-data}
    \vspace{-1.5em}
\end{figure}

\begin{table*}[t!]
\centering
\caption{\mdf{\textbf{Comparison with the state of the art on VidSTG DVOC validation set.} 
\gff{All models except Gemini \cite{comanici2025gemini} are finetuned on VidSTG. For the OVFormer baseline, we use the official checkpoint to finetune on VidSTG and use predicted category label names as captions.} "temp. agg." refers to including query temporal aggregation.}
}
    {
    \resizebox{0.7\textwidth}{!}{
    \begin{tabular}{l|c|ccc|c}
            \toprule
          \textbf{Method} & \textbf{Pretraining set}& \textbf{CapA} &  \textbf{DetA} & \textbf{AssA} & \textbf{CHOTA} \\
         \toprule
        \textcolor{lightgray}{Gemini-2.5-flash\cite{comanici2025gemini}}& \textcolor{lightgray}{-} & \textcolor{lightgray}{6.4} & \textcolor{lightgray}{0.9} & \textcolor{lightgray}{47.2} & \textcolor{lightgray}{6.6} \\
        \textcolor{lightgray}{OVFormer\cite{fang2024unified}} & \textcolor{lightgray}{LVIS + LV-VIS} &  \textcolor{lightgray}{12.8} & \textcolor{lightgray}{64.3} & \textcolor{lightgray}{50.1} & \textcolor{lightgray}{34.6} \\
         \midrule
         OW-VISCaptor 
         \citep{choudhuriow}
         & \multirow{2}*{COCO} & 43.9 & 60.1 & 54.0 & 53.0 \\
         \method{} (ours)  &  & 44.3 & 65.1 & 70.2 & 58.7 \\
         \midrule
         DVOC-DS 
         \citep{zhou2023dense} 
         & \multirow{2}*{COCO + VG + SMIT + AugCOCO} & 39.7 & {65.8} & {70.4} & {56.9} \\
         \method{} (ours)  & & 50.1 &  65.0 & 69.2 & 60.9 \\
        \midrule

         \method{} (ours) & COCO + LVIScap + LV-VIScap & {51.0} &  \textbf{66.8} & \textbf{71.0} & {62.3} \\
          + temp agg & COCO + LVIScap + LV-VIScap & \textbf{55.4} & \textbf{66.8} & \textbf{71.0} & \textbf{64.0} \\
         
    \bottomrule
    \end{tabular}
    }
    \label{tab:vidstg-comp}}
    \vspace{-1em}
\end{table*}

\begin{table}[t!]
\centering
\caption{\textbf{Comparison with state of the art on the VLN DVOC validation set.} Mask loss refers to using the segmentation masks in the detection loss. All models are finetuned on VLN. "temp. agg." refers to including query temporal aggregation.}
    \resizebox{0.47\textwidth}{!}{
    \begin{tabular}{l|c|ccc|c}
            \toprule
          \textbf{Method} & \textbf{\makecell[c]{Mask \\ loss}} & \textbf{CapA} & \textbf{DetA} & \textbf{AssA} & \textbf{CHOTA} \\
         \toprule
         DVOC-DS 
         {\cite{zhou2023dense}}
         & - &  17.7 & {44.3} & {89.5} & {41.3} \\
         
         \method{} (ours) & - &    {21.4} & {48.7} & {89.7} &  {45.4} \\
        \midrule
         \method{} (ours) & \ding{51} &  {22.9} & \textbf{50.1} & \textbf{92.7} & {47.4} \\
         + temp agg & \ding{51} & \textbf{23.4} & \textbf{50.1} & \textbf{92.7} & \textbf{47.7} \\
    \bottomrule
    \end{tabular}
    }
    \label{tab:vln-comp}
    \vspace{-1.5em}
\end{table}

\subsection{Results} \label{sec:results}

\subsubsection{Benefits of training on \idata{} and \vdata{}} \label{sec:benefits}
We first study the impact of integrating our generated datasets for DVOC training in Table \ref{tab:lvviscap}.
We train \method{} on our synthetic video set, \vdata{}, and progressively add \idata{} image pretraining. 
Results are reported on the \vdata{} validation set. 
First, we observe that \method{} achieves strong DVOC results with training only on \idata{} or \vdata{}, demonstrating the effectiveness of our architecture.
Importantly, combining both \idata{} and \vdata{} leads to best results, showing the benefit of both our generated datasets.
 Moreover, we observe that our \method{} is robust to the choice of the visual backbone. Note that the AssA score depends on the detections, hence a worse DetA score with lower recall but higher precision can make the tracking easier: this explains the higher AssA score from the model without LV-VIScap tuning.  \mdf{In Fig. \ref{fig:scaling-data}, we show that the CapA performance is logarithmically correlated with the quantity of training captions, suggesting that generating more data with our approach might bring further improvements.}

\FloatBarrier

\subsubsection{Comparison with the state of the art}\label{sec:sota}
\vspace{-0.1cm}
We compare \method{} to state-of-the-art DVOC methods following the standard evaluation protocol on three existing benchmarks : VidSTG in Table \ref{tab:vidstg-comp},  VLN in Table \ref{tab:vln-comp} and BenSMOT in Table \ref{tab:bensmot-comp}. 
DVOC-DS \citep{zhou2023dense} reports results without pretraining, and with their disjoint training strategy, while OW-VISCaptor \citep{choudhuriow} leverages Mask2former pretrained on COCO for instance segmentation.

\mdf{In Table \ref{tab:vidstg-comp}, we include results with the same pretraining data as these methods and show the impact of using our data instead.} 
Pretraining \method{} on COCO yields better detection and tracking compared to OW-VISCaptor but comparable captioning performance due to the similar captioning head design. Including our data leads to an improvement on all metrics and especially captioning with a 6.7 CapA increase (15\% relative improvement). When pretraining on the disjoint DVOC-DS set, we observe a substantial gain in the captioning metric due to the model design. Moreover, we show that using our pretraining sets results in a further performance improvement, while additionally allowing a unified, much faster training (2032 GPU hours \citep{zhou2023dense} vs 208 for our approach). Moreover, our proposed approach can output segmentation masks unlike the other methods.

We include two additional baselines : (i) \gff{We prompt Gemini-2.5-flash \cite{comanici2025gemini} with the video frames and instruct it to generate the (trajectory, caption) pair for each object. Results are retrieved from textual answers with some tolerance on the output format.} 
and (ii) 
a state of the art OV-VIS model, OVFormer \cite{fang2024unified}, 
by using predicted category label names as predicted captions. 
\gff{Gemini \cite{comanici2025gemini} is unable to provide correctly formatted answers in $13.6\%$ of the videos in the VidSTG validation set, and struggles to provide meaningful object regions, often generating near-stationary bounding boxes. This occasionally hinders any localized captioning capacity of the model, resulting in near 0 scores for both DetA and CapA.} 
\gff{On the other hand, OVFormer \cite{fang2024unified} is able to provide meaningful object regions, but lacks captioning ability: using predicted category labels as captions can retrieve some semantic information but misses context, relevant actions and states of the described entities. Moreover, OVformer struggles to track objects due to the length of VidSTG videos.}
\gff{Overall, these results further show the gap between understanding and localization,  highlighting the benefits of our unified DVOC approach.}
\begin{table}[t!]
\centering
\caption{\textbf{Comparison on the BenSMOT validation set.} CapA, and thus CHOTA are not reported on this dataset. All models are finetuned on BenSMOT. "temp. agg." refers to including query temporal aggregation.}
    \resizebox{0.8\linewidth}{!}{
    \begin{tabular}{l|cc|c}
        \toprule
        \textbf{Method} & \textbf{DetA} & \textbf{AssA} & \textbf{CIDEr}  \\
        \toprule
        SMOTer 
        { \citep{li2024beyond} }
        & 80.8 & 73.7 & 8.7 \\
        DVOC-DS 
        {\cite{zhou2023dense}}
        & 90.8 & \textbf{89.6} & 25.4\\
        \midrule
        \method{} (ours) & \textbf{91.6} & 87.5 & {39.9}\\
        + temp agg & \textbf{91.6} & 87.5 & \textbf{42.6} \\
        \bottomrule
    \end{tabular}}
    \label{tab:bensmot-comp}
    \vspace{-1em}
\end{table}

We further evaluate \method{} on VLN in Table \ref{tab:vln-comp} and BensMOT in Table \ref{tab:bensmot-comp}.
On both benchmarks, our approach improves  the detection while the tracking remains competitive. Most important, the captioning metrics improve by a large margin (+5.2 CapA on VLN, +14.7 CIDEr on BenSMOT compared to state of the art). Additionally, our method is able to jointly segment objects, which further improves the performance as shown in Table \ref{tab:vln-comp}. 

Across all benchmarks, we show that including temporal aggregation further improves captioning performance by effectively merging information from multiple video clips, permitting the description of temporally extended actions. 
We observe only a marginal improvement on VLN, likely due to the short video lengths of this dataset.
Because we do not modify the detection and tracking models when adding temporal aggregation, the DetA and AssA scores are unchanged.

\begin{table*}[t]
\centering
\begin{minipage}[t]{0.48\textwidth}
    \centering
    \captionof{table}{\mdf{\textbf{Automatic vs manual annotations for evaluation on a subset from LV-VIScap validation} with \camver{100} videos and \camver{476} objects trajectories ; "\textbf{automatic}" and "\textbf{manual}" annotations stands for synthetic or human annotated captions. All models are trained on \idata{} and tuned on our \vdata{} training set.}}
    \label{tab:manual_ann_bias}
    \resizebox{\textwidth}{!}{
    \begin{tabular}{c|c|ccc|c}
        \toprule
        \textbf{\makecell[c]{Annotation  \\ type}} & \textbf{\makecell[c]{LVIScap \\ captions}} & \textbf{CapA} & \textbf{DetA} & \textbf{AssA} & \textbf{CHOTA} \\
         \toprule
         
         \multirow{2}*{\textbf{automatic}} & - & 33.6 & 51.6 & 88.8 & 53.6 \\
         & \ding{51} & \textbf{41.4} & \textbf{52.1} & \textbf{90.2} & \textbf{58.0}\\
         \midrule
         \multirow{2}*{\textbf{manual}} & - & 22.9 & 51.6 & 88.8 & 47.1 \\
         & \ding{51} & \textbf{31.3} & \textbf{52.1} & \textbf{90.2} & \textbf{52.8} \\
    \bottomrule
    \end{tabular}}
    
\end{minipage}
\hfill
\begin{minipage}[t]{0.48\textwidth}
    \centering
    \captionof{table}{\textbf{Impact of temporal aggregation on VidSTG validation} (with finetuning). All methods are pretrained on COCO + LVIScap + LV-VIScap. \mdf{Multi-clip results are aggregated with weighted mean, except for $\dagger$ (arithmetic mean).}}\label{tab:vidstg-temp-agg-abl}
\resizebox{0.75\textwidth}{!}{
\begin{tabular}{c|c|cc} \toprule
            
          \textbf{num clips}& \textbf{clip selection} & \textbf{CapA} & \textbf{CHOTA} \\
         \toprule
          1 & best score & {51.0} &  {62.3} \\
           1 & middle frame & 46.9 &  60.6 \\
        4 & uniform$^{\dagger}$ & 49.1 &  61.5 \\
           4 & {uniform} & 51.6 &  62.6 \\
          8 &uniform & 52.7 &  63.0 \\
           16 & uniform & 53.8 &  63.4 \\
           32  &  uniform & \textbf{55.4} &  \textbf{64.0} \\
         
    \bottomrule \end{tabular}}
\end{minipage}%
\vspace{-1.5em}
\end{table*}

\begin{table}[t]
\centering
    \caption{\gf{\textbf{Impact of captioning on Video Instance Segmentation performance on LV-VIScap.}
    We train \method{} respectively with, and without enabling the captioning loss on our LVIScap and LV-VIScap datasets. }
    }
    \resizebox{0.2\textwidth}{!}{
    \begin{tabular}{c|cc}
        \toprule
          {\textbf{Captioning loss}} &   {\textbf{$\textrm{mAP}$}} 
          \\
         \toprule
          - & 31.7  \\
          \ding{51} & \textbf{34.2}  \\
    \bottomrule
    \end{tabular}
    \label{tab:ap_comparison}}
\vspace{-1.5em}
\end{table}

\begin{figure}[b!]
    \centering
    \vspace{-1em}
        \includegraphics[width=\linewidth]{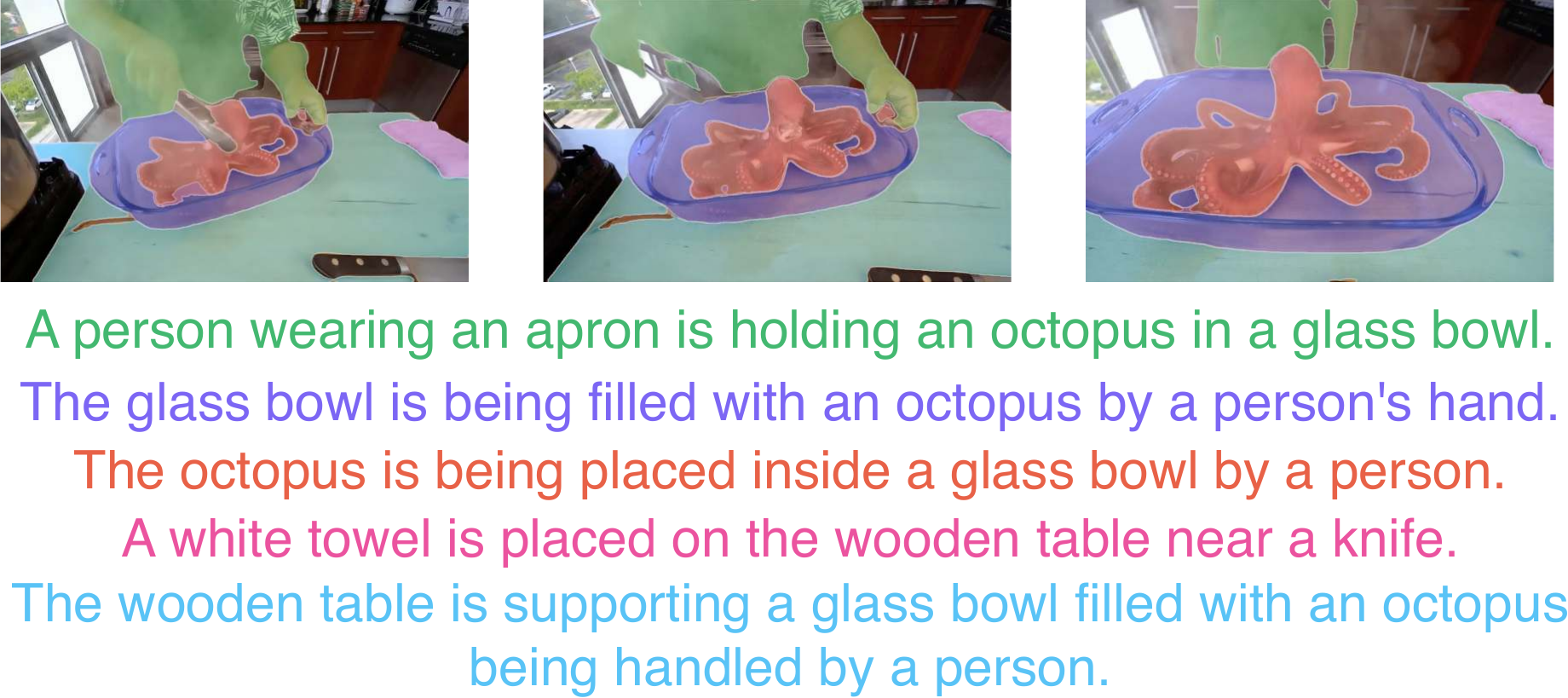}
        \vspace{-1.5em}
    \caption{Qualitative example obtained with \method{} on the LV-VIS dataset.}
    \label{fig:qualitative_res_lvvis_main}
\end{figure}

\subsubsection{Additional ablations} 
\label{sec:additional}

\mdf{\textbf{Annotation bias.} 
To evaluate the bias introduced by evaluating on \vdata{} synthetic captions (see Table \ref{tab:lvviscap}), we annotate a representative subset of our \vdata{} datasets by hand and compare the impact of our \idata{} captions when evaluating \method{} on \emph{automatic} vs \emph{manual} data.}
\mdf{Our subset includes \camver{100} videos from the \vdata{} validation set with \camver{476} object trajectories annotated by hand. 
The results are presented in Table \ref{tab:manual_ann_bias}. 
We observe that both the evaluation on automatic and manual data show a comparable improvement on the CapA and CHOTA metric when using \idata{} captions for training. 
This result confirms the importance of our synthetic captions for DVOC performance, and further shows that the bias introduced by evaluating on synthetic \vdata{} captions is marginal.}

\noindent \textbf{Temporal aggregation.}
\mdf{We show the impact of temporal aggregation on DVOC performance in Table~\ref{tab:vidstg-temp-agg-abl}.
Increasing the number of clips for aggregation consistently improves the captioning scores at the expense of higher training cost. Due to memory constraints we train with a maximum of 32 clips for aggregation on VidSTG.
Weighting the clips for aggregation using the detection scores yields a clear improvement over the arithmetic mean. We also observe that performing captioning based on the single clip with best score performs relatively well, which highlights the limits of the complexity of the actions observable in the current benchmarks, e.g. VidSTG~\citep{zhang2020does}}.

\noindent \textbf{Captioning loss.} \gf{We study the impact of the captioning loss on video instance segmentation (VIS) in Table \ref{tab:ap_comparison}, and observe that including our synthetic captions in the training improves $mAP$ by a considerable margin. This improvement indicates that the captioning supervision yields a rich training signal that is propagated to the object queries. Moreover, this result hints at potential benefits of the DVOC task for  large-vocabulary segmentation methods \cite{Wang_2023_ICCV}.}

\noindent \textbf{Qualitative example.} \gf{In Figure \ref{fig:qualitative_res_lvvis_main}, we show a qualitative sample produced by our \method{} on the LV-VIS dataset. \method{} effectively learns to jointly detect, track, segment and caption object trajectories in a video.}

\noindent \textbf{Additional results.}  We show additional qualitative examples in Supplementary~\ref{app-sec:qualitative1}, and discuss failure cases and limitations of our method in Section~\ref{app-sec:failure-limitation3}. 
Prompt ablation and the impact of the tracking module are presented in Section~\ref{app-sec:ablations} and further implementation details in Section \ref{app-sec:details}.

\section{Conclusion}
We propose an approach to generate synthetic object-level captions using a state-of-the-art VLM and extend the LVIS and LV-VIS datasets with synthetic captions. 
We use the resulting \idata{} and \vdata{} datasets to train \method{}, a DVOC model that can simultaneously detect, segment, track, and caption objects throughout a video.
With finetuning, \method{} achieves state-of-the-art performance on the VidSTG, VLN and BenSMOT benchmarks, \mdf{while extending the DVOC task to segmentation masks.}

\section*{Acknowledgements}

This work was granted access to the HPC resources of IDRIS under the allocation AD011014323R2 made by GENCI. It was funded in part by the French government under management of Agence Nationale de la Recherche as part of the “France 2030" program, reference ANR-23-IACL-0008 (PR[AI]RIE-PSAI project) and ANR project VideoPredict ANR-21-FAI1-0002-01.
Cordelia Schmid would like to acknowledge the support by the Körber European Science Prize.

{
    \small
    \bibliographystyle{ieeenat_fullname}
    \bibliography{main}
}
\clearpage
\setcounter{page}{1}
\twocolumn[{%
\renewcommand\twocolumn[1][]{#1}%
\vspace{1em}
\maketitlesupplementary
\centering
\vspace{1.5em}
\includegraphics[width=0.95\linewidth]{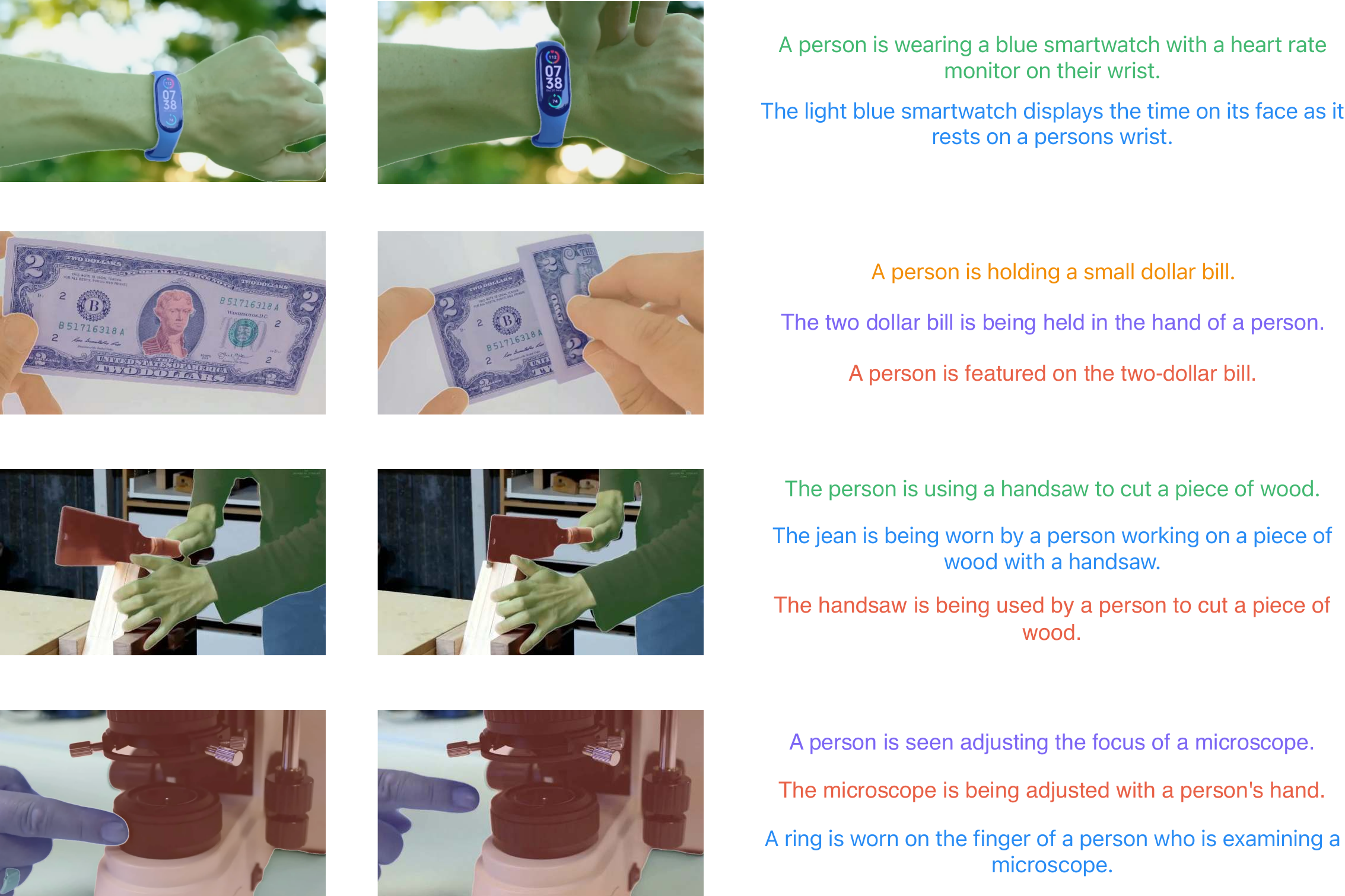}
\captionof{figure}{More qualitative examples obtained with \method{} on our \vdata{} dataset.}
\vspace{2em}
\label{fig:teaser-supp-mat}
}]

In this supplementary material, we show qualitative results from our method in Section \ref{app-sec:qualitative1}, present additional ablations in Section \ref{app-sec:ablations}, discuss failure cases and limitations in Section \ref{app-sec:failure-limitation3}, and share more details about the datasets, the method, and the implementation in Section \ref{app-sec:details}.

\section{Qualitative results} 
\label{app-sec:qualitative1}

\begin{figure*}[tp]
    \centering
        \vspace{1em}\includegraphics[width=0.82\linewidth]{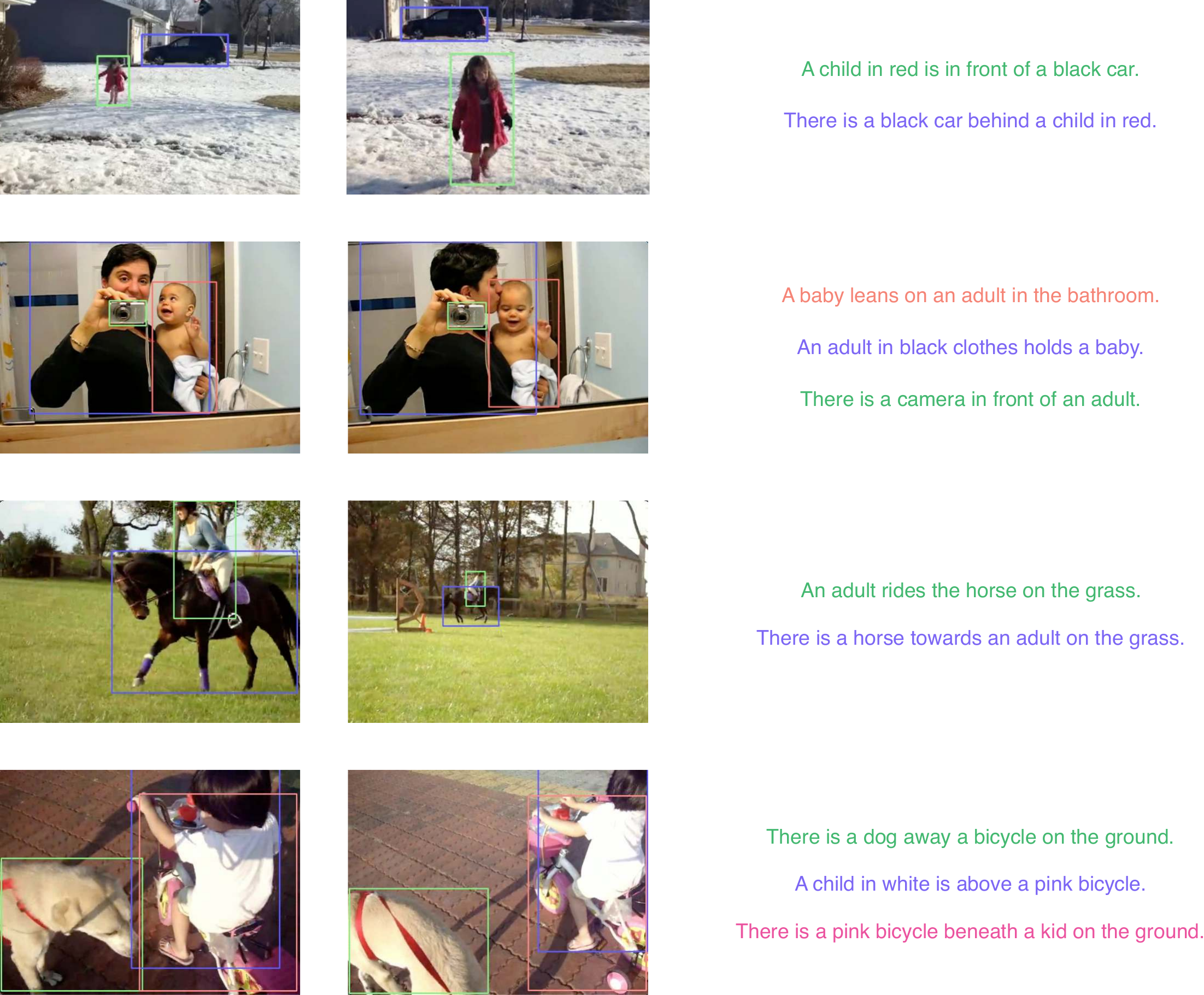}
    \caption{Qualitative examples, obtained with \method{} on the VidSTG dataset.}
    \label{fig:qualitative_res_vidstg}
\end{figure*}

We show qualitative DVOC results from \method{} on the LV-VIS \citep{Wang_2023_ICCV} dataset with (mask, caption) pairs predictions, and on the VidSTG dataset \citep{zhang2020does} with (box, caption) pairs in Figures \ref{fig:teaser-supp-mat} and \ref{fig:qualitative_res_vidstg} respectively.  
These examples show that \method{} has learned to predict captions that focus on the localized objects while integrating high-level scene understanding. 

On LV-VIS (Fig. \ref{fig:teaser-supp-mat}), \method{} is able to produce descriptive captions for each objects including related context, even in the case of a scene including a high number of objects. We note that, when tuned on VidSTG (see Fig. \ref{fig:qualitative_res_vidstg}), \method{} produces less informative and less descriptive captions. This is due to the VidSTG annotation captions being designed for grounding rather than for captioning or DVOC, and thus being only little descriptive or informative, and overlooking to the global context. In contrast, when trained on \idata{} and \vdata{}, \method{} visually generates much richer and accurate descriptions, further highlighting the value of our synthetic captions. \\
Overall, \method{} effectively learns to jointly segment, detect, track and caption object trajectories.

\begin{table*}[t]
\caption{\textbf{Impact of the prompting strategy on caption quality}. Scores are given by a human evaluator from 0 to 2 (incorrect, partially correct, or correct) on a subset from the LV-VIS validation set, and brought to 0-100 range. For the mask visual prompt experiments, we use our best prompt with either the object's bounding boxes or center point coordinates as a localization cue in the text prompt.}
\centering
\resizebox{0.7\linewidth}{!}{
    \begin{tabular}{c|l|c|ccc}
\toprule
\multirow{2}*{\textbf{Visual prompt}} & \multirow{2}*{\textbf{Prompting method}} &  \multirow{2}*{\textbf{Average rating}} & \multicolumn{3}{c}{\textbf{Rating percentage}} \\
\cmidrule(lr){4-6}
& & & \textbf{0} & \textbf{1} & \textbf{2} \\
\toprule
\multirow{7}*{{bounding boxes}} & single frame &  26.8 & 66.3 & 13.9 & 19.9 \\
& + multiple frames &  27.1 & 68.7 & 8.4 & 22.9 \\
& + detailed instructions &  29.5  & 65.0 & 10.8 & 24.10 \\
& + category labels & 80.7 & 10.8 & 16.9 & 72.3 \\
& + bounding box coordinates &  83.1 &  9.6 & 14.5 & 75.9  \\
 & + bounding box area &  \underline{84.3} & 7.8 & 15.7 & 76.5 \\
& + few shot examples &  \textbf{85.1} &  9.1 & 11.5 & 79.4 \\
\midrule 
\multirow{2}*{{mask boundaries}}  & center point coordinates &  75.9 & 17.5 & 13.2 & 69.3 \\
 &  bounding box coordinates & 77.1 & 15.7 & 14.5 & 69.9 \\
\bottomrule
\end{tabular}
\label{tab-app:prompt-ablation-appendix}}
\vspace{-0.2em}
\end{table*}

\section{Additional ablations}
\label{app-sec:ablations}
\vspace{-0.4em}

\textbf{Prompting strategy.}
In Table \ref{tab-app:prompt-ablation-appendix}, we show the distribution of ratings given by the human annotator depending on the prompting strategy. Using the box visual prompt yields a better focus on the queried object and more correct object captions. Importantly, giving the category labels in the prompt helps the model to generate more accurate object captions. Overall, the rate of correct captions with the best prompt indicates good quality for our synthetic object captions.

\begin{table}[h]
\centering
\caption{\mdf{\textbf{Impact of the inference clip on LV-VIScap.} All models are trained on LVIScap then LV-VIScap training set.
}}
\resizebox{0.4\linewidth}{!}{
\begin{tabular}{c|c}
        \toprule
      \textbf{Clip length} &  \textbf{mAP} \\
     \toprule
    3 & 26.0 \\
    4 & 26.1 \\
    5 & \underline{26.6} \\
    6 & 26.3 \\
    7 & \textbf{26.8} \\
\bottomrule
\end{tabular}
\label{tab:clip-len}}
\end{table}

\noindent\textbf{Clip Length.} We study the impact of the clip length on Video Instance Segmentation (VIS) performance in Table \ref{tab:clip-len}, and observe that increasing clip length leads to slightly better segmentation results. Overall the impact on performance is relatively small. For LV-VIS experiments, we use a clip length of $T=5$.

\begin{table}[h]
\centering
\caption{
\textbf{Impact of the tracking module on VidSTG DVOC.} All models are trained on COCO, \idata{} and \vdata{}, and finetuned on VidSTG using temporal aggregation with 8 clips.
}
    \resizebox{\linewidth}{!}{
    \begin{tabular}{c|ccc|c}
        \toprule
        \textbf{Tracking method} &  \textbf{CapA} & \textbf{DetA} & \textbf{AssA} & \textbf{CHOTA} \\
         \toprule
           OVFormer module \citep{fang2024unified} & 51.9 & \textbf{67.0} & 58.2 & 58.7 \\
           
          Top-K enhanced module \citep{zhu2024clip} & \textbf{52.7} & {66.8 }& \textbf{71.0} & \textbf{63.0} \\

    \bottomrule
    \end{tabular}
    \label{tab:tracking_ablation}}
\end{table}

\noindent\textbf{Tracking module.}
In Table \ref{tab:tracking_ablation}, we show that the top-k enhanced tracking~\citep{zhu2024clip} is important for tracking objects effectively on the challenging VidSTG dataset \citep{zhang2020does}, as seen in the AssA, CapA and CHOTA scores. \\
We attribute this difference to the numerous objects that disappear for a significant number of frames in the long videos of VidSTG. 
The top-K approach uses a memory bank of tracked queries that helps keeping track of these entities, while they are lost using the $i$ to $i+1$ tracking from OVFormer \citep{fang2024unified}. \\

\begin{table}[h]
    \centering
    \captionof{table}{\textbf{Zero-shot DVOC performance} on the VidSTG validation set. 
    Our model is pre-trained on COCO, \idata{} and \vdata{}, while DVOC-DS is pre-trained on COCO, VG, SMiT, Aug-COCO.
    }
    \label{rebuttal:zero-shot-vidstg}
    \resizebox{0.48\textwidth}{!}{
    \begin{tabular}{l|ccc|c}
        \toprule
        \textbf{Method} & \textbf{CapA} & \textbf{DetA} & \textbf{AssA} & \textbf{CHOTA} \\
         \toprule
         DVOC-DS~\cite{zhou2023dense} & 9.8 & \textbf{51.4 }& 59.6 &  {31.1} \\
         \method{} (ours) & \textbf{10.4} & 50.2 & \textbf{71.3} & \textbf{33.4}\\
    \bottomrule
    \end{tabular}}
\end{table}

\noindent\textbf{Generalization to Out-of-Distribution data.} \camver{We report zero-shot performance on the VidSTG validation set in Table~\ref{rebuttal:zero-shot-vidstg}, and compare with DVOC-DS~\cite{zhou2023dense}. We observe that our model achieves better performance in the zero-shot setting for captioning and tracking. The DVOC-DS performs slightly better for detection, which can be explained by their larger pretraining set for detection. }

\begin{table}[h]
\centering
    \caption{\textbf{Inference speed comparison} on a subset of the VidSTG validation set using a single A100 GPU.}
    \label{tab:fps-comp}
    \begin{tabular}{l|c}
    \toprule
         \textbf{Method} & \textbf{FPS} \\
         \toprule
         OW-VisCapTor~\cite{choudhuriow} & 6.4 \\
         \method{} (ours)& \textbf{7.4} \\
    \bottomrule
    \end{tabular}
    
\end{table}

\noindent\textbf{Inference speed.} 
\camver{We compare the inference speed of our model to OW-VisCapTor~\citep{choudhuriow} in Table~\ref{tab:fps-comp}. The OW-VisCapTor numbers were provided by the authors for a subset of the VidSTG videos and  a A100 hardware setup. We run our approach on the same subset and hardware for fair comparison.}

\section{Failure Cases and limitations}
\label{app-sec:failure-limitation3}
\subsection{Failure cases}
\label{app-subssec:C-1-failure}

\begin{figure*}[htb]
    \centering
    \includegraphics[width=0.96\linewidth]{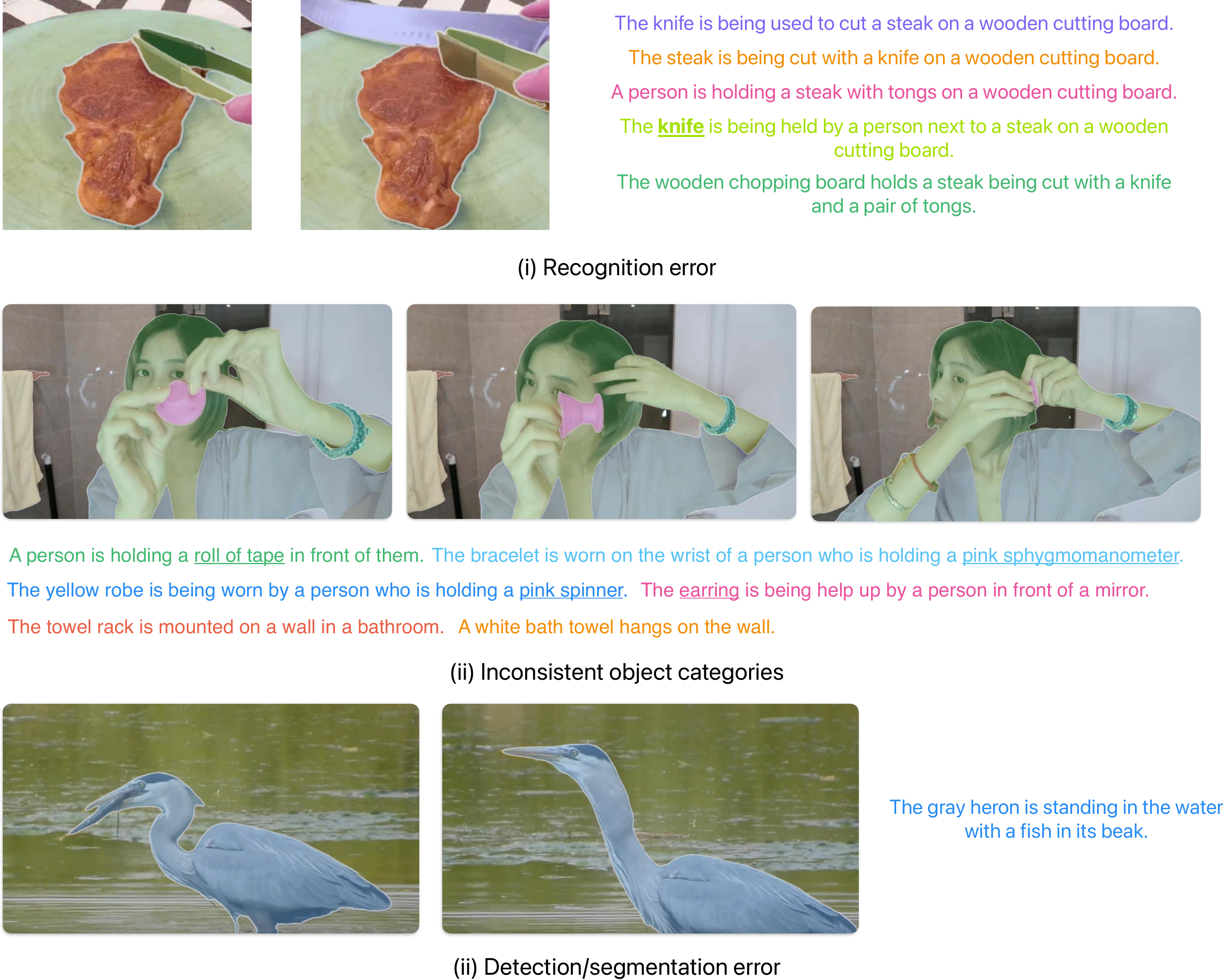}
    \vspace{-0.2cm}
    \caption{Some DVOC failure cases of \method{} observed on the LV-VIS dataset.}
    \label{fig:failure-case}
\end{figure*}

We observe 3 main types of failure cases for our approach and illustrate them in Figure \ref{fig:failure-case}: \\
\textbf{(i) Recognition error}: in the case of ambiguous context, blurred instance or rare categories, \method{} might fail to recognize the object it is describing, sometimes leading to a wrong denomination (e.g.~here, a rare "pair of tongs" is incorrectly denominated as "knife"). \\ 
\textbf{(ii) Inconsistent captions} : in similar situations, 
the captions produced by \method{} can be inconsistent when referring to the same object in different captions. \\
\textbf{(iii) Detection/segmentation error} : In case of complex movement, appearance change or occlusion, \method{} sometimes fails to detect, segment, or track objects, leading to missing captions (e.g.~here, the fish is not detected in the beak of the heron, and thus has no associated caption).

\subsection{Limitations}
\label{app-subssec:C-1-limit}

\label{sec:limitations}
While our approach outperforms the state-of-the-art in dense video object captioning, there is still room for improving localization and captioning.
Localization sometimes fails, in particular for small objects. 
Furthermore, the automatically generated captions are, in some cases, 
too generic, and can mix up two objects of the same class in the video.
Future work could investigate different automatic captioning techniques for DVOC, for example based on an approach such as Ref-SAV\citep{sa2va}, which generates captions in multiple steps to separate appearance from motion description.
\mdf{Eventually, objects in the videos often perform a single or few actions, and we believe that it is important for future works to build benchmarks with more complex object interactions, for instance with multiple action segments.}

\vspace{-0.1em}
\section{Additional details}
\label{app-sec:details}

\vspace{-0.1em}
\subsection{Prompting strategy details}
\label{app-subsec:prompt-strat}

\begin{figure}[h!]
    \centering
    \includegraphics[width=\linewidth]{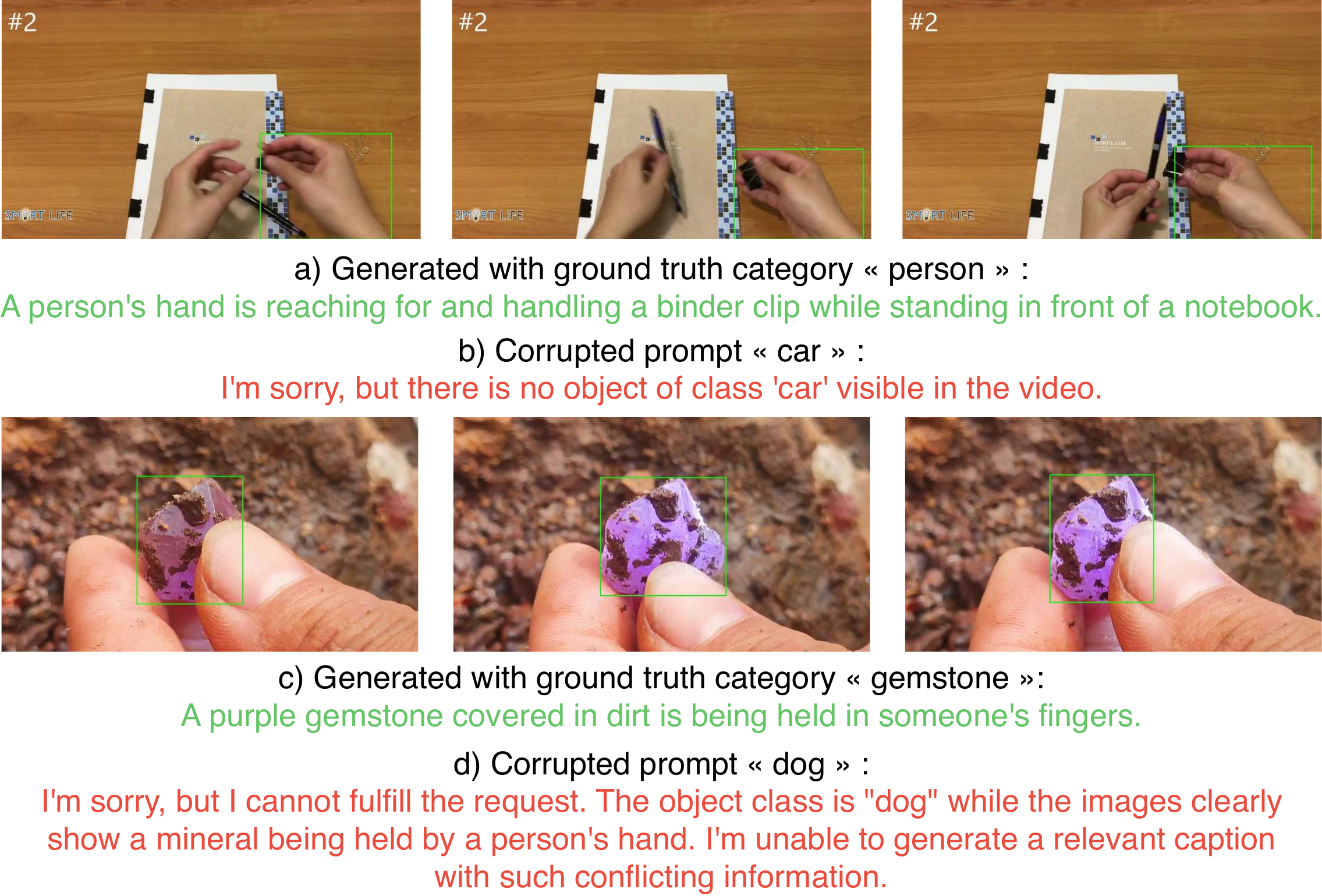}
    \caption{{\textbf{Qualitative examples} generated with our synthetic annotation approach using ground truth label and corrupted labels in the prompt.}
    }
    \label{fig:suppmat-prompt-qualitative}
\end{figure}

\camver{The bounding boxes used for the visual prompt are rendered with a very thin line (2px), which we observe is sufficiently thin to not impact recognition. 
Figure~\ref{fig:suppmat-prompt-qualitative} shows that our prompting strategy effectively guides Gemini to attend to the image and focus on the object of interest, resulting in additional visual details being included in the generated captions. In the top example we can observe that Gemini refines the label "person" to "a person's hand" and describes the interaction with the object and environment. In the bottom example the label "gemstone" is refined by the color purple and the fact that it is covered in dirt.} 

\camver{The category labels given in the prompt are human-annotated ground truth. 
Nevertheless, we experimented with incorrect labels and 
Fig.~\ref{fig:suppmat-prompt-qualitative}~(red) shows that Gemini correctly identifies the resulting inconsistency.} 

\begin{figure*}[t]
    \centering
    \includegraphics[width=\linewidth]{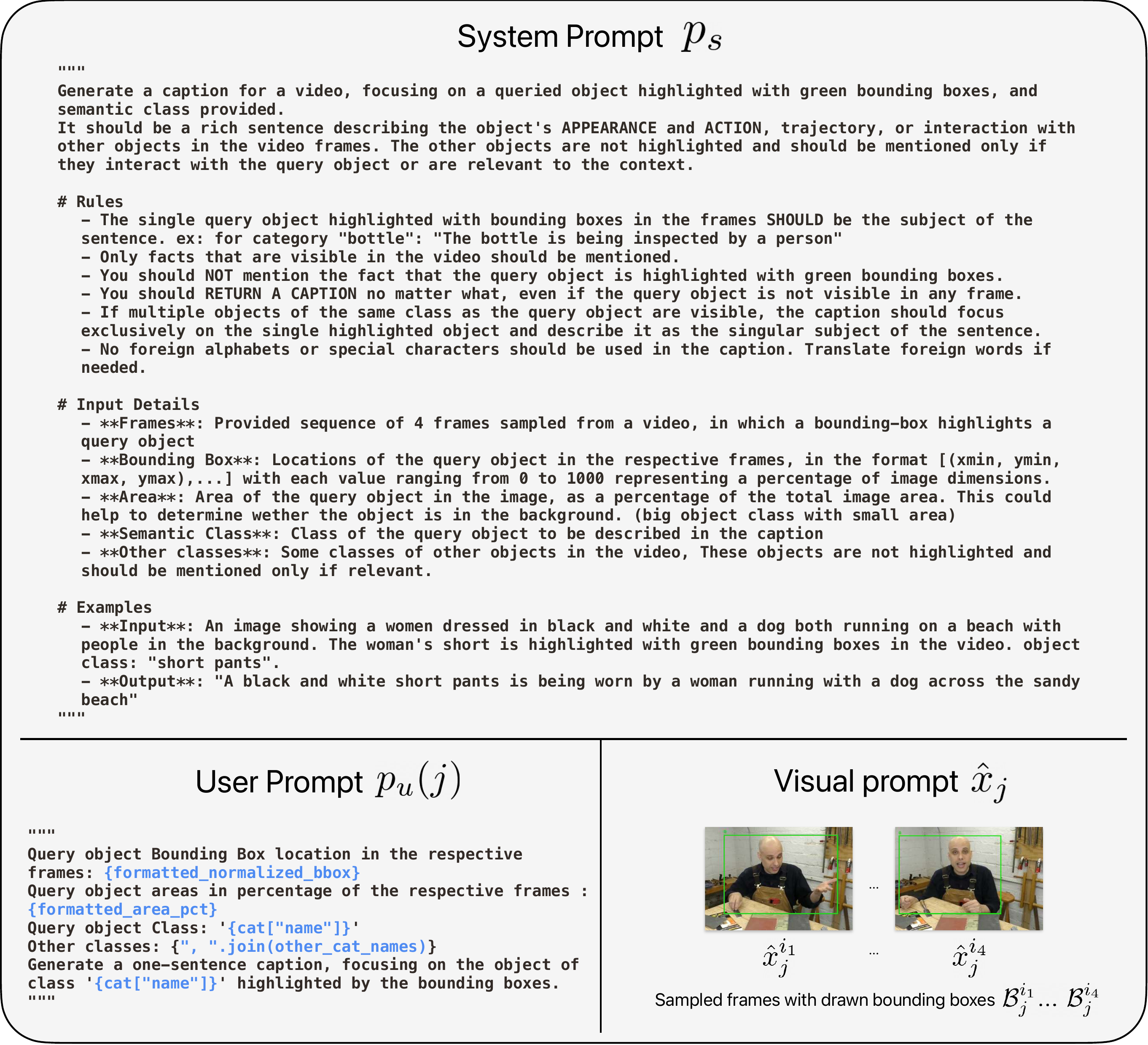}
    \caption{Prompt template used to generate synthetic object captions from video segmentation annotations for the LV-VIS dataset \cite{Wang_2023_ICCV}. 
    The system prompt $p_s$ contains general instructions, while user prompt and visual prompt $p_u(j)$ and $\hat{x}_j$ are formatted with information from each annotation.}
    \label{fig:prompts}
    \vspace{-1.2em}
\end{figure*}

The full prompt template used to generate our \idata{} and \vdata{} datasets is illustrated in Figure \ref{fig:prompts}. 
For a video $x$ with $N$ objects, we prompt the VLM $N$ times, and for each object $j$ the prompt is composed in three parts: 
(i) the static system prompt $p_s$ gives general instructions for object-level caption generation, practical rules, prompting format and an example,
(ii) the user prompt $p_u(j)$ depends on the example and contains textual annotation information to help the model describe objects and interactions accurately. 
These information notably contain target object positions, areas, category, and the categories of other objects in the scene,
(iii) the visual prompt $\hat{x}_j$ consists of $4$ sampled frames with drawn bounding boxes for object $j$.

\vspace{-0.2em}
\subsection{Dataset details}
\vspace{-0.1em}
\label{app-sec:datasets}
\textbf{LVIS} \citep{gupta2019lvis} is a large-vocabulary image instance segmentation dataset with a long-tail distribution of $1203$ annotated categories, for a total of over $2.2$ million annotations in $164$k natural images. 
The dataset is split in a training set with $100$k images and $1.2$M annotations, a validation set with $19$k images and $244$k annotations, and two test sets with $19$k images each. \\

\noindent\textbf{LV-VIS} \citep{Wang_2023_ICCV} is a recent large-vocabulary video instance segmentation (VIS) benchmark. 
It comprises $4,828$ videos with over $26$k video segmentation masks from 1,196 object categories, with an average of over $5.4$ objects per video. 
The data is split into a training set of $3,076$ videos and $16$k video-level annotations, a validation set of $837$ videos and $3.7$k annotations, and a test set with $908$ videos.\\
\textbf{\idata{}} and \textbf{\vdata{}} denote our extensions of LVIS and LV-VIS (see Section \ref{sec:datagen}).
\idata{} extends LVIS with a total of $1,488,354$ synthetic captions, including $1,244,271$ training annotations and $244,083$ validation annotations.
\vdata{} includes a total of $19,717$ synthetic captions for $16,017$ training and $3,700$ validation annotations. 
Note that in the absence of annotations on the test sets of LVIS and LV-VIS, we only extend the training and validation sets with captions, and use the validation set for evaluation. \\
\textbf{VidSTG}\citep{zhang2020does} is a spatio-temporal video grounding dataset, containing $6,924$ videos with $44,808$ exhaustive trajectories annotations over $80$ categories, as well as object sentence descriptions (for some objects and some timestamps only), which serve as queries for grounding. 
\citet{zhou2023dense} repurposed the dataset for DVOC by using queries as captions, and by excluding annotations without captions during evaluation. 
Following \citet{zhou2023dense}, we sample 200 frames uniformly across each video for both training and evaluation. 
Overall, the repurposed VidSTG training set counts $28,275$ object trajectories, with $15,182$ object captions. 
The validation set, used for DVOC evaluation, includes  $602$ videos with $1,644$ captions. \\
\textbf{Video Localized Narratives (VLN)} extends existing datasets with "narrations of actors actions" in videos. We use the subset from the UVO dataset, which contains 3 sparsely annotated frames with non exhaustive captions for a total of $5,136$ training and $2,451$ validation videos. \\
\noindent\textbf{BenSMOT} contains manually collected annotations of bounding box trajectories and associated captions, focusing exclusively on humans in videos. It includes an average of $2.2$ instances per video, and counts $2,284$ videos for training and $1,008$ for evaluation.

\subsection{More Implementation details}
\label{app-sec:implementation}

The visual backbone is initialized with weights pretrained on ImageNet-21K \cite{deng2009imagenet} following OVFormer \cite{fang2024unified}, and the Mask2Former \cite{Cheng_2022_CVPR} weights are trained from scratch. 
The OVFormer classifier uses a frozen CLIP ViT-B/32 \cite{radford2021learning} encoder. 
The captioning head is initialized with weights from BLIP-2 \cite{li2023blip} with frozen LLM OPT-2.7B \cite{zhang2022opt} following Chouduri et al. \cite{choudhuriow}. 

For all experiments except LV-VIS tuning, we first train the segmentation/detection model, then freeze it and  tune the captioning head. Respectively for LVIS/VidSTG/LV-VIS we train for $440k/40k/22k$ for the first stage and $5k/2k/2k$ for the second stage. When tuning pretrained models on VidSTG/\mdf{VLN/BenSMOT, we train the two stages for $(15k,2k)/(15k,500)/(15k,2k) $ steps}, whereas for LV-VIS, we end-to-end tune the model for $2k$ steps. 
Experiments are run with a batch size of $8$, except when using LVIS+COCO and LV-VIS where we use a batch size of $4$ \mdf{and for video-level tuning of the captioning head where we use a batch-size of 1. Experiments on LV-VIS are end-to-end trainings with clip-level supervision only. For VidSTG/VLN/BenSMOT experiments we use video-level tuning for captioning with temporal aggregation, with $T_{\textrm{agg}}=32/8/8$ respectively.}
For all experiments we train the model with a clips of size $T=2$, and at inference use $T=5/1/1/1$, $T_{\textrm{match}}=1/100/20/40$, $K_{\textrm{match}}=1/7/5/7$ for LV-VIS/VidSTG/VLN/BenSMOT experiments respectively. For the largest dataset (COCO + LVIS) the optimization takes 2 days on 4 H100 GPUs.

Following OVFormer \cite{fang2024unified}, for all datasets we use an AdamW optimizer and the step learning rate schedule, with an initial learning rate of 0.0001 and a weight decay of 0.05, and apply a 0.1 learning rate multiplier to the backbone. We decay the learning rate at 0.9 and 0.95 fractions of the total number of training steps by a factor of 10. 
For respectively image/video datasets, we resize the shortest edge of the image to 800/480 for SwinB and 800/360 for ResNet for training and inference.

\end{document}